\documentclass[11pt,letterpaper]{article}
\usepackage[margin=1in]{geometry}
\usepackage[T1]{fontenc}
\usepackage{lmodern}
\usepackage{microtype}
\usepackage{mathtools,amssymb}
\usepackage{graphicx}
\usepackage{booktabs,makecell,tabularx}
\usepackage[font=small,labelfont=bf]{caption}
\usepackage{needspace}
\usepackage{cite}
\usepackage{xurl}
\usepackage[hidelinks]{hyperref}

\makeatletter
\newcommand{\useexistingpdfgroup}{%
  \ifdefined\Gread@transgroupfalse
    \let\Gread@transgrouptrue\Gread@transgroupfalse
  \fi}
\makeatother

\hypersetup{
  pdftitle={What Does Tool Ablation Measure? A Text-to-SQL Agent Study},
  pdfauthor={Saksonita Khoeurn, Ha Jeong Kim, Yeji Yoon, and Sereiwathna Ros}
}
\newcommand{\csql}{\texttt{convertSmqToSql}}
\newcommand{\gme}{\texttt{getModelDataElements}}
\newcommand{\sqlex}{\texttt{SqlExecution.execute}}
\newcommand{\dsch}{\texttt{describeSchema}}
\newcommand{\lcol}{\texttt{lookupColumns}}
\newcommand{\icat}{\texttt{inspectCatalog}}
\newcommand{\SR}{\mathrm{SR}}
\newcommand{\Sm}{S_{2m}}
\newcommand{\pp}{\,pp}

\newcommand{\rateci}[2]{\makecell{#1\\{\footnotesize[#2]}}}
\newcommand{\effectp}[2]{\makecell[t]{#1\\{\footnotesize $p=#2$}}}
\newcommand{\effectci}[3]{%
  \makecell[t]{#1\\{\footnotesize[#2]}\\{\footnotesize $p=#3$}}}

\title{What Does Tool Ablation Measure?\\A Text-to-SQL Agent Study}
\author{Saksonita~Khoeurn\thanks{S.~Khoeurn is with the Department of Management
Information Systems, Chungbuk National University, Cheongju, Republic of Korea
(e-mail: \href{mailto:saksonita@chungbuk.ac.kr}{\nolinkurl{saksonita@chungbuk.ac.kr}}).
She is the corresponding author.},
Ha Jeong~Kim\thanks{H.~J.~Kim
(\href{mailto:hajkim@daquv.com}{\nolinkurl{hajkim@daquv.com}}) and Y.~Yoon
(\href{mailto:philoyyj@daquv.com}{\nolinkurl{philoyyj@daquv.com}}) are with
DAQUV Corp., Republic of Korea.},
Yeji~Yoon\footnotemark[\value{footnote}],
and Sereiwathna~Ros\thanks{S.~Ros is with the Department of Computer Science,
Chungbuk National University, Cheongju, Republic of Korea.}}
\date{}

\begin{document}

\maketitle

\begin{abstract}
Tool ablation (removing a tool and measuring the drop in task success) is
a common approach to evaluating an LLM agent's dependence on a tool. However,
many tools also serve as the agent's view of its environment, so the ablation
effect conflates what the tool does with what it allows the agent to observe.
We investigate this distinction through three contributions. First, we
introduce a read-only substitute that removes the tool's operation while
restoring access to the information it exposes, decomposing the ablation
effect into operational and perceptual contributions. Second, we
construct an experimental testbed for a semantic-layer text-to-SQL agent whose
SQL translator is its only direct source of physical schema names, comprising
six automatically generated conditions on 332 Spider~2.0-Lite tasks across
SQLite and BigQuery, three LLM backbones, and a renamed-schema control, with
every condition run three times. Third,
we demonstrate that conventional ablation can obscure the sources of tool
dependence. On BigQuery with Gemini~3.1~Pro, a small, statistically
non-significant ablation effect ($+6.6$ percentage points, $p=0.07$) masks two
significant opposing effects: a negative translator contribution ($-18.3$
points) and a positive schema-access contribution ($+24.9$ points). When
schemas are renamed, the success rate of Gemini without the translator falls
from $76.8\%$ to $22.2\%$, suggesting substantial translator dependence. Yet a
read-only lookup returning semantic-to-physical mappings outperforms the
translator across all eight experimental settings, significantly in six. These
findings reveal that tool dependence does not necessarily imply operational
usefulness and suggest that semantic-layer agents benefit from exposing
physical schema mappings independently of query compilation.
\end{abstract}

\begin{quote}
\end{quote}

\section{Introduction}
\label{sec:intro}
Large language model (LLM) agents act through tools, and
data systems are increasingly among those tools
\cite{fernandez2023llms,biswal2025tag}. Much of agent evaluation asks which
tools matter. The standard answer comes
from ablation \cite{meyes2019ablation}: remove one tool, keep everything else
fixed, and read the change in task success as the tool's contribution
\cite{liu2025agentbenchevaluatingllmsagents,NEURIPS2024_877b4068}.
Tool-use benchmarks score tools one at a time
\cite{patil2025bfcl,chen-etal-2024-eval,ICLR2024_28e50ee5} or compare agents
with and without a tool, and the resulting numbers inform which tools a
production system keeps. In data systems the question arises directly for
LLM-based natural-language interfaces to databases
\cite{liu2025survey,hong2025next}, whose tools decide what the model can see
of the schema.

Interpreting a change in task success requires knowing what the agent loses
when a tool is removed. In
the usual agent loop every tool call returns an observation
\cite{yao2023react}, so a schema tool in a database agent
\cite{ICLR2025_46c10f6c,zhou2024dbot}, a page reader in a web agent
\cite{ICLR2024_4410c071} and a file viewer in a coding agent
\cite{yang2024sweagent} each perform an action and, at the same time, show
the agent part of its environment. Removing such a tool removes both. The
ablation drop is then a valid measure of how much the agent depends on the
tool \emph{as a package}, but it cannot say whether the agent needed the
tool's operation or only the view the tool happened to provide. Work on
removal-based explanation makes the same point for model inputs and
internals: what a removal measures depends on what is put in place of the
removed part, and different replacements can lead to different conclusions
\cite{covert2021explaining,hooker2019benchmark,zhang2024activation}. The
distinction matters in practice: a tool whose value is mostly perceptual can
be replaced by a cheaper read-only channel, while a tool whose operation
itself helps cannot. How an agent's interface presents the environment is
known to change its success \cite{yang2024sweagent}, and for text-to-SQL,
giving the model the schema directly can stand in for schema-linking steps
\cite{chung2025long}.

\begin{figure}[t]
\centering
\useexistingpdfgroup
\includegraphics[width=\linewidth]{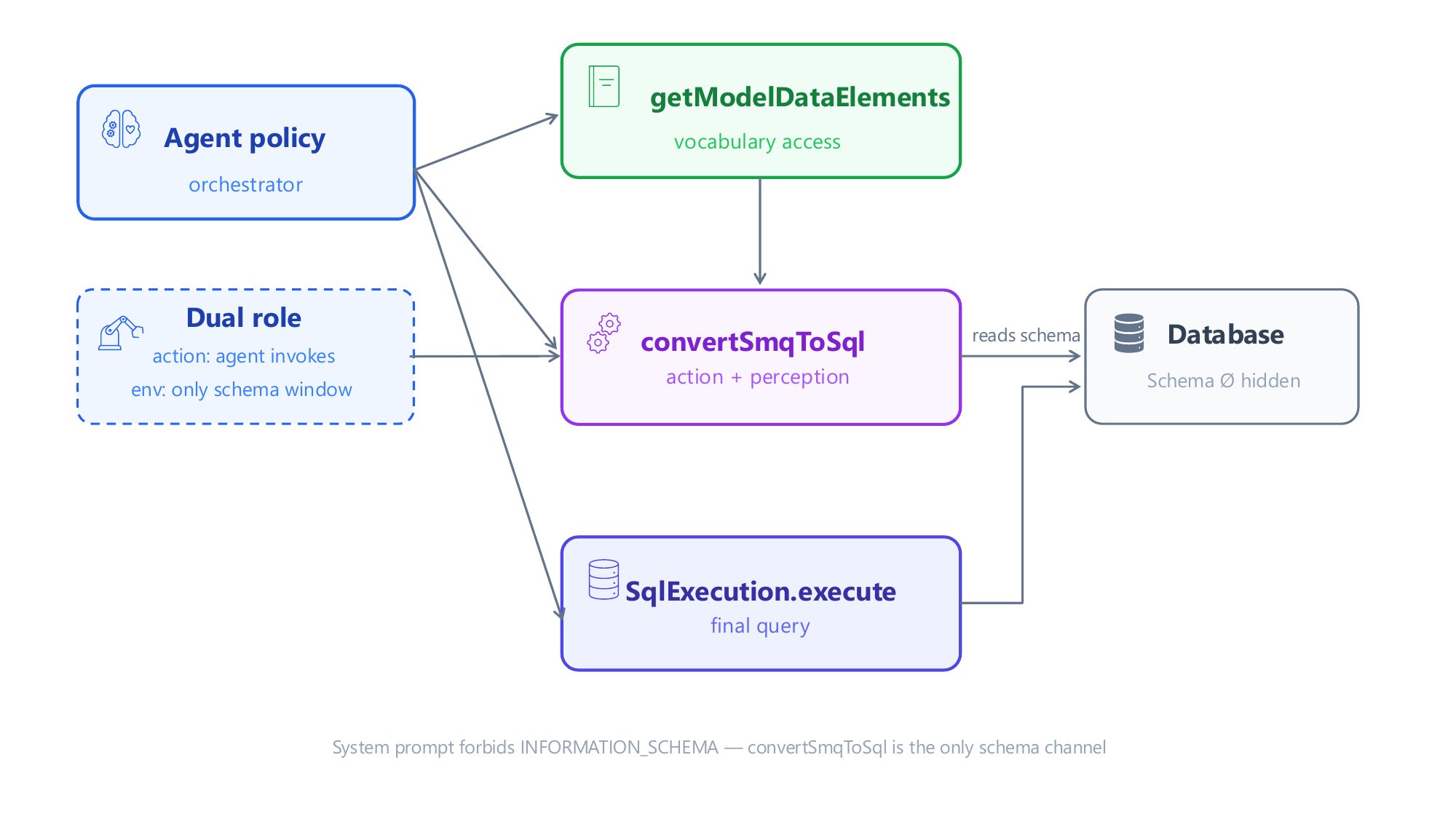}
\caption{The agent studied in this paper. The agent policy calls three
tools: \gme{} returns the semantic layer's vocabulary but no physical names;
\csql{} compiles a semantic query into SQL against the database; \sqlex{}
runs the final query. Because the prompt forbids catalog
(\texttt{INFORMATION\_SCHEMA}) queries, the physical schema is hidden, and
\csql{} plays a dual role (dashed box): an action the agent invokes and the
agent's only window onto physical table and column names.}
\label{fig:arch}
\end{figure}

Fig.~\ref{fig:arch} shows the case we study, a semantic-layer text-to-SQL
agent \cite{kim2026semantic}. A semantic layer places a business vocabulary of models, metrics and
dimensions between users and physical tables. The agent reads that
vocabulary through \gme{}, which shows no physical names. To answer a
question it calls \csql{}, which turns a structured query over the vocabulary
into SQL by reading the database schema, and it submits its final query
through \sqlex{}. Since catalog queries are forbidden, the SQL returned by
\csql{} is the only place the agent sees physical table and column names.
\csql{} is therefore both an action (compiling SQL) and a perceptual channel
(revealing the schema). If we remove it, the agent loses the compiler and
its view of the schema at once, and the ablation drop cannot say which loss
matters.

The two losses can be distinguished by giving the agent access to the schema
after removing the compiler. A \emph{perception-preserving substitute} $S$
provides the information previously exposed by the tool through a read-only
channel. For our agent, this amounts to replacing \csql{} with a lookup that
returns the physical table and column behind each semantic element, without
compiling anything. With the full system $A$ and the ablation $C$, the
ablation effect splits exactly:
\begin{equation}
\underbrace{\SR(A)-\SR(C)}_{\text{total dependence}}
=\underbrace{\SR(A)-\SR(S)}_{\text{tool's own contribution}}
+\underbrace{\SR(S)-\SR(C)}_{\text{restoring the view}} ,
\label{eq:decomp-intro}
\end{equation}
where $\SR$ is the task success rate. The split is only as good as the
substitute's informational match to $A$, so we define that match
operationally, as the physical schema elements the agent can observe before
it answers, and measure it rather than assert it.

For the agent in Fig.~\ref{fig:arch}, we compare six conditions (the full system, an augmented
reference, three substitutes and the ablation) on 332 tasks from the SQLite
and BigQuery portions of Spider~2.0-Lite~\cite{ICLR2025_46c10f6c}, with
three backbones of different strength (Gemini~3.1~Pro, Qwen3.5-27B, and
Qwen3.5-9B on SQLite only) and a SQLite control in which every physical name
is replaced by an opaque code, so that semantic names no longer reveal
physical ones. Every condition is run three times, and every comparison is
paired at the task level.

Where removing the translator produces a small, non-significant change in
success, the lookup reveals two large effects of opposite sign. With the
same schema information the translator lowers success, while the view it
provides raises it. When renamed physical names make that view
scarce, ablation calls the translator indispensable, yet a read-only lookup
that returns the same information still outperforms it. Across two SQL
dialects and three backbones, the lookup scores higher than the translator in
every setting, significantly in six of eight: what the agent needs from
\csql{} is its view of the schema, not its compiler.

\noindent Our main experimental findings and contributions are summarized as
follows:
\begin{itemize}
\item \textbf{Conventional ablation can mask opposing effects.} By introducing
a read-only substitute that restores schema access without query compilation,
we find that a small, statistically non-significant ablation effect can
conceal substantial negative and positive contributions associated with the
translator interface and schema access, respectively.
\item \textbf{Tool dependence changes with information availability.} Through
experiments on 332 Spider~2.0-Lite tasks across two SQL dialects, three LLM
backbones, and a renamed-schema control, we show that conventional ablation is
highly sensitive to the agent's ability to infer physical schema identifiers.
\item \textbf{Tool dependence does not necessarily imply operational
usefulness.} Across all eight experimental settings, a read-only mapping
lookup outperforms the translator, significantly in six. This finding
suggests that semantic-layer agents can benefit from exposing
semantic-to-physical mappings independently of query compilation.
\end{itemize}

\section{Related Work}
\label{sec:related}

\subsection{Evaluating Tool Use in LLM Agents}
Tool-use benchmarks progressed from whether a model can call the right API
\cite{NEURIPS2024_e4c61f57,ICLR2024_28e50ee5} to scoring individual calls
\cite{patil2025bfcl}, decomposing tool use into sub-skills
\cite{chen-etal-2024-eval}, multi-turn and stateful interaction
\cite{NEURIPS2024_877b4068,lu2025toolsandboxstatefulconversationalinteractive,yao2024taubenchbenchmarktoolagentuserinteraction}
and full environments \cite{ICLR2024_4410c071,liu2025agentbenchevaluatingllmsagents}.
Attribution to a single tool in these settings is usually done by ablation or
by scoring that tool's calls in isolation. Both assume that the rest of the
agent's environment is unchanged when the tool is removed or scored alone.
ToolSandbox \cite{lu2025toolsandboxstatefulconversationalinteractive} models
state that one tool call passes to the next; our concern is the
complementary case in which the tool \emph{is} the agent's access to that
state. This case is common in data-system agents, whose tools both act on the
system and expose its state, as in database diagnosis with monitoring tools
\cite{zhou2024dbot}. In the data-management literature, NL2SQL360
\cite{li2024dawn} shows that a single accuracy number hides large differences
across scenarios; we make the analogous point for a single ablation number. In contrast to these benchmarks, which ask how well an agent uses a
tool, this paper asks what removing a tool actually takes away.

\subsection{Ablation and Its Interpretation}
Ablation is borrowed from lesion studies of neural networks
\cite{meyes2019ablation}, where removing a component is understood to change
the network's downstream computation, not only its output. In agents the same
holds at the level of information: a removed tool changes what the agent
observes. Causal mediation analysis
\cite{pearl2001direct,vanderweele2015explanation} separates a total effect
into a direct effect and an effect through a mediator. We use its vocabulary
to read our decomposition, with perception as the mediator and the substitute
as the device that holds it fixed. Unlike mediation analysis in observational
data, the mediator here is set by design, and we verify its value by
measurement rather than by modelling assumptions.
Table~\ref{tab:related} summarizes how our approach differs from these
existing ways of attributing success to a tool.

\begin{table}[t]
\caption{Comparison With Existing Approaches to Tool Attribution.
\checkmark: supported; $\times$: not supported; --: not applicable
(a general framework, not specific to tools).\label{tab:related}}
\centering
\small
\begin{tabular}{@{}lcccc@{}}
\toprule
Approach & \makecell{Single-tool\\attribution} & \makecell{Accounts for\\what tool shows} & \makecell{Information\\measured} & \makecell{Signed\\components} \\
\midrule
Tool ablation \cite{liu2025agentbenchevaluatingllmsagents} & \checkmark & $\times$ & $\times$ & $\times$ \\
Per-call scoring \cite{patil2025bfcl,chen-etal-2024-eval} & \checkmark & $\times$ & $\times$ & $\times$ \\
Mediation analysis \cite{pearl2001direct,vanderweele2015explanation} & $\times$ & \checkmark & -- & \checkmark \\
\textbf{Ours} & \checkmark & \checkmark & \checkmark & \checkmark \\
\bottomrule
\end{tabular}
\end{table}

\subsection{Text-to-SQL Agents and Semantic Layers}
Text-to-SQL has been surveyed from the deep-learning era
\cite{katsogiannis2023survey,kim2020natural} to the LLM era
\cite{liu2025survey,hong2025next}. Benchmarks such as Spider \cite{yu2018spider}
and BIRD \cite{li2023bird} give the model the schema; pipelines such as
DIN-SQL \cite{pourreza2023din}, DAIL-SQL \cite{gao2024dail} and CodeS
\cite{li2024codes} build on that access, and their accuracy depends on how the
schema is represented in the prompt \cite{gao2024dail}. Schema linking decides
how much of the schema the model sees \cite{luoma2026schema}, long-context
models can instead be given all of it \cite{chung2025long}, and how natural the
schema identifiers are affects accuracy as well \cite{luoma2025snails}.
Spider~2.0 \cite{ICLR2025_46c10f6c} moves to enterprise databases where
schemas are large and partly undocumented and agents must explore, and
practitioners report that enterprise NL2SQL remains far from solved for the
same reasons \cite{floratou2024nl2sql}. Semantic layers place a business
vocabulary (metrics, dimensions, joins) between users and physical tables,
defining reusable metrics over them \cite{hyde2024measures}. The idea predates
LLMs: ontology-driven interfaces such as ATHENA \cite{saha2016athena} map a
domain ontology to SQL over the physical schema, and for LLMs a
knowledge-graph layer over an enterprise SQL database has been reported to
raise question-answering accuracy \cite{sequeda2024benchmark}. Such
evaluations compare systems with and without the layer; they do not ask
whether the gain comes from the layer's operations or from what it reveals
about the schema. An agent that works only through the layer sees physical names only
when a tool translates layer elements into SQL, which makes the translator
both an action and a perceptual channel. This is the case this paper studies.

\section{Problem Formulation and Decomposition}
\label{sec:framework}

\subsection{Setting}
An agent configuration is a tuple (backbone, tool set, prompt, environment,
task set). For a task $t$ and condition $X$, let $y_X(t)\in[0,1]$ be the
task's success averaged over independent runs, and $\SR(X)$ its mean over
tasks. Success is exact agreement of the agent's result table with the gold
result (Section~\ref{subsec:scoring}).

Let $\tau$ be the tool under study. Three conditions are needed.
\begin{itemize}
\item $A$: the full configuration, $\tau$ available.
\item $C$: $\tau$ removed and nothing put in its place (ablation).
\item $S$: $\tau$ removed, and the information that $\tau$ exposed about the
environment returned through a read-only channel that performs none of
$\tau$'s operation (substitute).
\end{itemize}

\subsection{The Decomposition}
Equation~(\ref{eq:decomp-intro}) holds for any $S$. It becomes informative
when $S$ exposes the same environment information as $A$. Write $P$ for what
the agent can observe and $T$ for whether $\tau$'s operation is available.
$A$ has $T=1$ and $P=p_A$; $C$ has $T=0$ and $P=p_C$; a matched substitute
has $T=0$ and $P=p_A$. Then $A-S$ compares configurations that differ in
$T$ with $P$ held at $p_A$, the analogue of a controlled direct effect
\cite{pearl2001direct}, and $S-C$ compares configurations that differ only in
$P$ with $T=0$, the effect of restoring perception. We do not claim causal
identification beyond the design: a substitute necessarily differs from $A$ in
interface (prompt text, number of turns, format of returned information), and
we measure the most important of these differences
(Section~\ref{subsec:info}) instead of assuming them away.

When the available substitute exposes \emph{more} than $A$, a second
substitute that exposes less closes the gap. With a richer substitute
$S_2$ and a matched one $\Sm$,
\begin{equation}
A-C=\underbrace{(A-\Sm)}_{\text{translator, info.\ equal}}
+\underbrace{(\Sm-S_2)}_{\text{extra information}}
+\underbrace{(S_2-C)}_{\text{schema access}} .
\label{eq:decomp3}
\end{equation}

\subsection{Informational Equivalence, Measured}
\label{subsec:equiv}
That a substitute ``provides the same information'' is usually a design
intention. We replace it with a measurement. For a task $t$ and a
run of condition $X$, let $V_X(t)$ be the set of physical schema elements
(columns, identified by table and name) whose names appear, as whole words,
in what the agent observed before submitting its final query: the initial
prompt plus every non-execution tool response. Two summaries are reported:
the mean size $|V_X|$, and gold-column coverage, the share of columns used
by the gold query that lie in $V_X$. A substitute is \emph{information-matched}
to $A$ when both summaries agree within run-to-run variation. This definition
is computed automatically from logs and the database catalog (on BigQuery,
the semantic layer); it is
conservative (it counts names, not whether the agent understood them) and
specific to schema perception, but the same construction applies to any
environment whose relevant state can be enumerated.

\subsection{Reading the Components}
Ablation alone answers ``how much does this configuration depend on the
tool package?'' This is a legitimate question, and the right one for a
deployment decision about removing the tool with nothing in its place. The decomposition
answers two further questions: would a read-only channel do as well
($A-S$), and how much of the dependence is about seeing the environment
($S-C$)? The components can have opposite signs, in which case ablation can
be small while both components are large. We report every component with a
paired confidence interval. A one-sided substitution index such as
$\min(1,\SR(S)/\SR(A))$ would discard the sign that the decomposition
exposes, so we report the components themselves.

Table~\ref{tab:quadrants} gives the qualitative readings. The cell that
motivates this work is \emph{necessary, not unique}: ablation collapses,
so the tool looks indispensable, yet a substitute recovers or exceeds $A$.

\begin{table}[t]
\caption{Readings of the decomposition.\label{tab:quadrants}}
\centering
\small
\begin{tabularx}{\linewidth}{@{}l*{2}{>{\raggedright\arraybackslash}X}@{}}
\toprule
 & $S\gtrsim A$ (own contribution $\le 0$) & $S\ll A$ (own contribution $>0$) \\
\midrule
$S-C$ small & tool is weak & tool helps through its operation; view not needed \\
$S-C$ large & \textbf{necessary, not unique}: needed only as a view & tool needed for both operation and view \\
\bottomrule
\end{tabularx}
\end{table}

\section{Testbed}
\label{sec:testbed}
\begin{figure}[t]
\centering
\useexistingpdfgroup
\includegraphics[width=\linewidth]{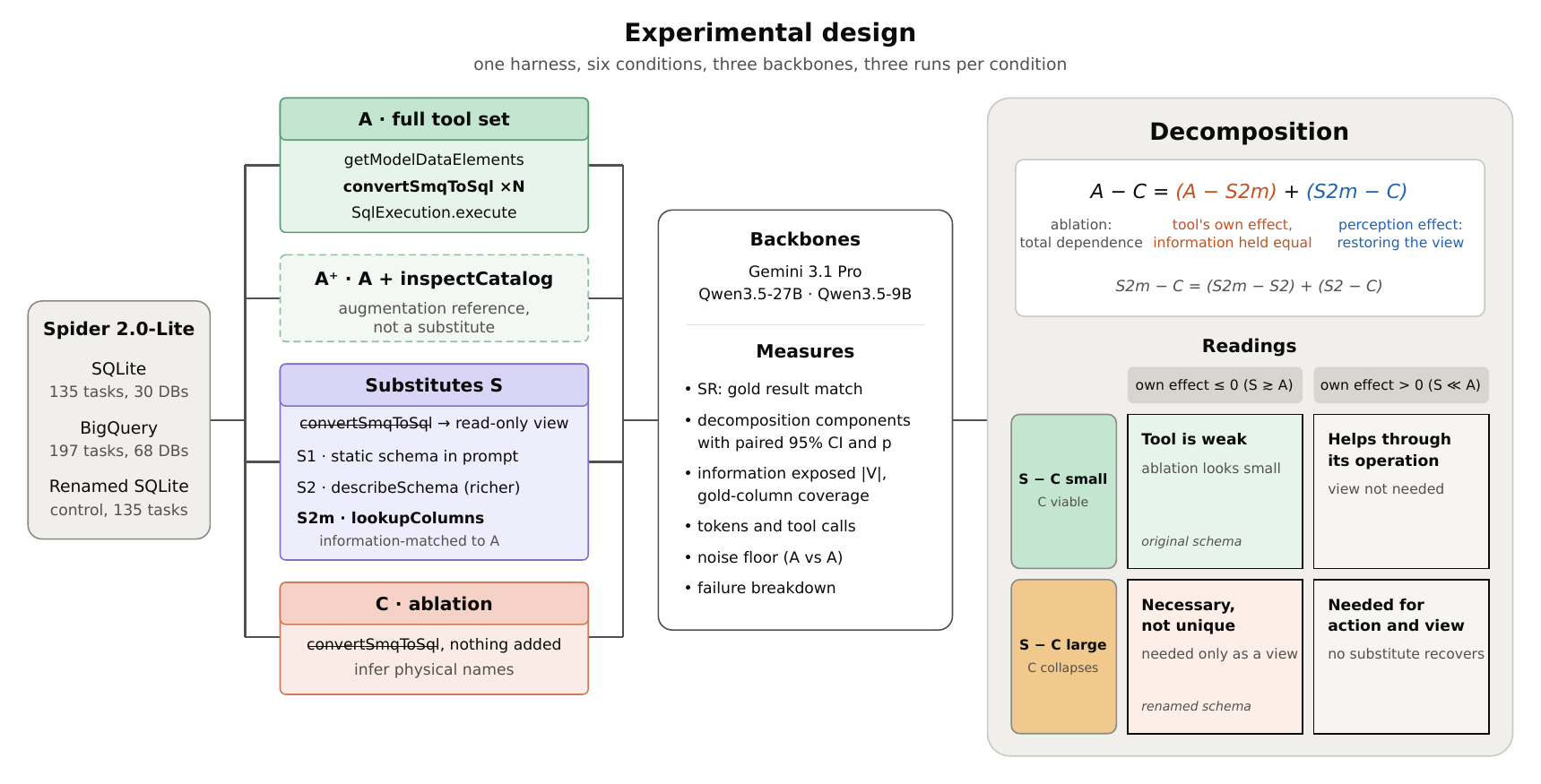}
\caption{Experimental design. Every task is run under six conditions on one
harness; the matched substitute $\Sm$ splits the ablation effect into the
translator's own effect and the perception effect, whose signs give the
readings on the right.}
\label{fig:design}
\end{figure}

Fig.~\ref{fig:design} gives an overview of the experimental design: the
tasks, the six conditions, the backbones and measures, and how the
decomposition of Section~\ref{sec:framework} is read from the results.

\subsection{The Semantic-Layer Agent}
The agent is QUVI \cite{kim2026semantic}, a semantic-layer text-to-SQL
agent built on a Java tool-calling harness. By February 2026, with Gemini~3 Pro, it
ranked first on all three tracks of the Spider~2.0 leaderboard (Snow, Lite
and DBT), reaching 94.15\% execution accuracy on
Spider~2.0-Snow~\cite{kim2026semantic,thelec2026daquv}. Its semantic layer, written in YAML, defines for each
database a set of semantic models (one per table), their dimensions and
measures with physical expressions, metrics, and join relations. In the full
configuration the agent has three tools:
\begin{itemize}
\item \gme{} returns the layer's vocabulary: model, metric and dimension
names with descriptions and the join graph, but no physical names;
\item \csql{} takes a structured semantic query (metrics, group-by
dimensions, filters, order, limit), compiles it with the layer's engine into
dialect SQL and returns that SQL with preview rows;
\item \sqlex{} runs the agent's final SQL and ends the episode.
\end{itemize}
The prompt forbids catalog queries. Because the engine joins only one hop and
cannot express every query, the agent typically calls \csql{} several times,
reads physical expressions from the returned SQL and writes the final query
itself. \csql{} is therefore both an action (compiling SQL) and the agent's
only view of physical names (Fig.~\ref{fig:arch}).

\subsection{Conditions}
All six conditions run on the same harness, backbone settings and turn cap,
and differ only in tool list and prompt (Table~\ref{tab:conditions}).

\begin{table}[t]
\caption{Conditions. All keep \gme{} and \sqlex{}.\label{tab:conditions}}
\centering
\small
\begin{tabularx}{\linewidth}{@{}l>{\raggedright\arraybackslash}X>{\raggedright\arraybackslash}p{0.29\linewidth}@{}}
\toprule
Cond. & What changes & Role \\
\midrule
$A$ & full tool set, \csql{} available & baseline \\
$A^{+}$ & $A$ plus read-only \icat{} & augmentation reference \\
$S_1$ & \csql{} removed; static schema of the database in the prompt & substitute, prompt-layout stress test \\
$S_2$ & \csql{} removed; \dsch{} returns every column (name, type, description) of the models asked for & substitute, richer than $A$ \\
$\Sm$ & \csql{} removed; \lcol{} returns, for each semantic element named, only its physical table and expression & \textbf{information-matched substitute} \\
$C$ & \csql{} removed, nothing added; prompt asks the agent to infer physical names from semantic ones & ablation \\
\bottomrule
\end{tabularx}
\end{table}

$\Sm$ is constructed to reveal exactly what \csql{} reveals about an
element (its physical table and expression) and nothing else: it composes
no SQL and runs nothing. $A^{+}$ is not a substitute; it adds a catalog tool
on top of $A$ and indicates whether more information helps the full
configuration. The static schema of $S_1$, the responses of \dsch{} and
\lcol{}, and the $C$ prompt are all generated by one script from the
semantic layer and the database catalog, so the substitutes need no manual
engineering per database.

\subsection{Tasks and Databases}
\noindent\textbf{SQLite.} All 135 SQLite tasks of Spider~2.0-Lite over 30
databases, each with a semantic layer.

\noindent\textbf{BigQuery.} The layers of the BigQuery databases had been
written against a Snowflake copy of the data. We ported them: 229 semantic
models were repointed to their BigQuery tables (16 had to be relocated to
another dataset), and 2,502 of 2,591 dimension and measure expressions were
rewritten from Snowflake to BigQuery SQL (timestamps stored as integers in the
copy, semi-structured paths, lateral flattening, time zones, string splitting,
then a dialect transpiler) and validated by a BigQuery dry run; the rest
reference tables that are no longer readable or columns absent from BigQuery.
Of the 205 BigQuery and GA tasks, 197 on 68 databases have a ported layer and
readable tables and form the task set. In $S_1$, $S_2$ and $\Sm$ tables are
shown fully qualified (\texttt{project.dataset.table}), as \csql{} writes
them; $C$ must infer the project and dataset as well, which matters for the
79 tasks whose layer spans more than one dataset.

\noindent\textbf{Engine coverage.} Compiling every metric of every layer
and dry-running it, 604 of 671 SQLite metrics and 369 of 436 BigQuery
metrics succeed. The failures are engine limits (one-hop joins, join keys
that must be plain columns, empty metrics) that affect only conditions that
use the engine, i.e.\ $A$ and $A^{+}$. We consider them part of the tool
under study and do not repair them.

\subsection{Renamed-Schema Control}
\label{subsec:renamed}
On Spider the physical names mirror the semantic names (dimension
\texttt{Album\_\_albumId} lives in column \texttt{AlbumId}), and public
schemas may be memorized. Both let $C$ guess physical names. Identifier
naturalness is known to affect LLM text-to-SQL accuracy
\cite{luoma2025snails}; our control removes it entirely. For the 30
SQLite databases we built copies in which every table and column has an
opaque name (\texttt{t\_8e107}, \texttt{c\_06e4c}), with data unchanged, and
rewrote the layers' physical expressions accordingly while keeping all
semantic names. Version~1 left the layers' descriptions untouched, so 213 of
2,312 descriptions still cite old names (``renamed schema with outdated
documentation''). Version~2, the main control, also rewrites names inside
descriptions.

\subsection{Backbones and Settings}
Gemini~3.1~Pro (preview) \cite{gemini31} runs through Vertex AI with thinking
disabled; Qwen3.5-27B (FP8) and Qwen3.5-9B \cite{qwen35} run on vLLM on two
NVIDIA L4 GPUs with native thinking off, a 64K context and 8,192 output
tokens. All runs use temperature 0.1 and a cap of 20 tool calls. Gemini
occasionally answers the text tool-call protocol with a native function call
(\texttt{MALFORMED\_FUNCTION\_CALL}); the harness resends the prompt up to
three times. Qwen3.5-27B often omits the final closing brace of a tool call
that carries a long SQL string; the harness parser closes unbalanced
brackets before parsing, for every condition (Section~\ref{subsec:failures}).

\subsection{Scoring and Statistics}
\label{subsec:scoring}
The agent's final SQL is re-executed and its result table compared with the
gold result(s) by the official Spider~2.0-Lite evaluator, which compares
results rather than SQL strings \cite{kim2020natural}. Every condition is
run three times: all six on Gemini, and $A$, $S_2$, $\Sm$ and $C$ on Qwen.
Per-task success
is averaged over runs. Differences between conditions come with a 95\%
task-level bootstrap interval \cite{efron1993bootstrap} and a two-sided paired
sign-flip permutation test (10,000 resamples each); single-run comparisons
also report McNemar's test \cite{mcnemar1947}. The noise floor is measured by
comparing repeated runs of $A$.

\Needspace{6\baselineskip}
\section{Experiments}
\label{sec:results}
We evaluate the decomposition on the testbed of Section~\ref{sec:testbed}
around three research questions:
\begin{itemize}
\item \textbf{RQ1.} Does the decomposition change the conclusion that
ablation alone would draw? (Gemini, SQLite and BigQuery, all six conditions.)
\item \textbf{RQ2.} Is the pattern a property of a strong backbone?
(Qwen3.5-27B and Qwen3.5-9B, a different family and two sizes.)
\item \textbf{RQ3.} Does it rest on physical names that mirror the semantic
layer or on memorized schemas? (Renamed schemas, all three backbones.)
\end{itemize}
Section~\ref{sec:analysis} then checks the informational match, explains
where the translator loses, tests prompt sensitivity, presents a case study,
and reports the noise floor and the cost.

\begin{table}[t]
\caption{Success rate (\%, mean of three runs) with 95\% task
bootstrap intervals for the Gemini runs.\label{tab:sr}}
\centering
\small
\setlength{\tabcolsep}{2.5pt}
\begin{tabular*}{\linewidth}{@{\extracolsep{\fill}}llcccccc@{}}
\toprule
Backbone & \makecell[l]{Database\\(tasks)} & $A$ & $A^{+}$ & $S_1$ & $S_2$ & $\Sm$ & $C$ \\
\midrule
Gemini 3.1 Pro & SQLite (135)
  & \rateci{80.0}{74.6, 84.9} & \rateci{79.8}{73.8, 85.2} & \rateci{84.0}{77.8, 89.4}
  & \rateci{86.4}{80.7, 91.4} & \rateci{\textbf{88.4}}{83.2, 93.1} & \rateci{76.8}{70.4, 83.0} \\
\addlinespace[2pt]
Gemini 3.1 Pro & BigQuery (197)
  & \rateci{64.8}{59.1, 70.1} & \rateci{75.8}{70.9, 80.4} & \rateci{76.6}{71.1, 82.2}
  & \rateci{78.2}{72.4, 83.2} & \rateci{\textbf{83.1}}{78.3, 88.0} & \rateci{58.2}{52.3, 64.3} \\
\addlinespace[2pt]
Qwen3.5-27B & SQLite (135) & 59.5 & -- & -- & 56.8 & \textbf{64.2} & 51.6 \\
Qwen3.5-27B & BigQuery (197) & 45.0 & -- & -- & 56.7 & \textbf{57.9} & 9.6 \\
Qwen3.5-9B & SQLite (135) & 15.6 & -- & -- & \textbf{26.9} & 20.7 & 16.8 \\
\bottomrule
\end{tabular*}
\end{table}

\begin{table}[t]
\caption{Decomposition of the ablation effect, Eq.~(\ref{eq:decomp3}).
Percentage points, mean over three runs; 95\% task bootstrap interval and paired
permutation $p$. The three components add up to $A-C$ (up to rounding).\label{tab:decomp}}
\centering
\small
\setlength{\tabcolsep}{3pt}
\begin{tabular*}{\linewidth}{@{\extracolsep{\fill}}llccccc@{}}
\toprule
Backbone & Schema
  & \makecell{$A-C$\\(ablation)}
  & \makecell{$A-\Sm$\\(translator)}
  & \makecell{$\Sm-S_2$\\(extra info)}
  & \makecell{$S_2-C$\\(schema access)}
  & \makecell{$\Sm-C$\\(perception)} \\
\midrule
Gemini & SQLite
  & \effectci{$+3.2$}{$-2.0$, $+8.6$}{.27}
  & \effectci{$\mathbf{-8.4}$}{$-12.6$, $-4.2$}{.0002}
  & \effectp{$+2.0$}{.39}
  & \effectp{$\mathbf{+9.6}$}{.003}
  & \effectci{$\mathbf{+11.6}$}{$+6.9$, $+16.5$}{.0001} \\
\addlinespace[3pt]
Gemini & BigQuery
  & \effectci{$+6.6$}{$-0.3$, $+13.5$}{.07}
  & \effectci{$\mathbf{-18.3}$}{$-23.0$, $-13.7$}{.0001}
  & \effectp{$\mathbf{+4.9}$}{.013}
  & \effectp{$\mathbf{+20.0}$}{.0001}
  & \effectci{$\mathbf{+24.9}$}{$+19.1$, $+30.6$}{.0001} \\
\addlinespace[3pt]
Qwen 27B & SQLite
  & \effectci{$\mathbf{+7.9}$}{$+1.2$, $+14.8$}{.03}
  & \effectci{$-4.7$}{$-11.1$, $+2.0$}{.18}
  & \effectp{$\mathbf{+7.4}$}{.04}
  & \effectp{$+5.2$}{.21}
  & \effectci{$\mathbf{+12.6}$}{$+6.4$, $+18.8$}{.0002} \\
\addlinespace[3pt]
Qwen 27B & BigQuery
  & \effectci{$\mathbf{+35.4}$}{$+28.6$, $+42.1$}{.0001}
  & \effectci{$\mathbf{-12.9}$}{$-19.1$, $-6.8$}{.0004}
  & \effectp{$+1.2$}{.70}
  & \effectp{$\mathbf{+47.0}$}{.0001}
  & \effectci{$\mathbf{+48.2}$}{$+41.3$, $+55.0$}{.0001} \\
\addlinespace[3pt]
Qwen 9B & SQLite
  & \effectci{$-1.2$}{$-7.4$, $+4.4$}{.75}
  & \effectci{$-5.2$}{$-10.1$, $-0.2$}{.06}
  & \effectp{$-6.2$}{.08}
  & \effectp{$\mathbf{+10.1}$}{.002}
  & \effectci{$+4.0$}{$-0.7$, $+8.6$}{.13} \\
\midrule
Gemini & renamed v2
  & \effectp{$\mathbf{+56.8}$}{.0001}
  & \effectci{$\mathbf{-8.4}$}{$-13.1$, $-4.0$}{.0008}
  & \effectp{$+3.0$}{.16}
  & \effectp{$\mathbf{+62.2}$}{.0001}
  & \effectp{$\mathbf{+65.2}$}{.0001} \\
\addlinespace[3pt]
Qwen 27B & renamed v2
  & \effectp{$\mathbf{+32.6}$}{.0001}
  & \effectci{$\mathbf{-13.3}$}{$-21.0$, $-5.7$}{.0008}
  & \effectp{$+3.2$}{.38}
  & \effectp{$\mathbf{+42.7}$}{.0001}
  & \effectp{$\mathbf{+45.9}$}{.0001} \\
\addlinespace[3pt]
Qwen 9B & renamed v2
  & \effectp{$+4.2$}{.08}
  & \effectci{$\mathbf{-7.4}$}{$-13.1$, $-1.7$}{.016}
  & \effectp{$-2.7$}{.51}
  & \effectp{$\mathbf{+14.3}$}{.0001}
  & \effectp{$\mathbf{+11.6}$}{.0004} \\
\bottomrule
\end{tabular*}

\vspace{2pt}
\raggedright\small Bold: $p<0.05$.
\end{table}

\subsection{Main Results (RQ1): Ablation Hides Two Opposing Effects}
\label{subsec:rq1}
\begin{figure}[t]
\centering
\includegraphics[width=0.68\linewidth]{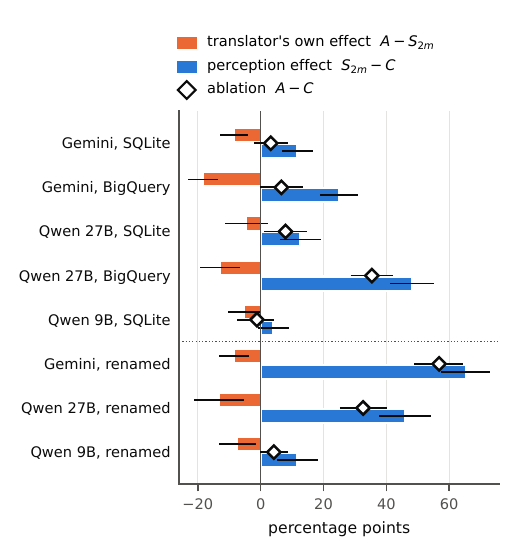}
\caption{Ablation hides two opposing effects. Diamonds: the ablation effect
$A-C$ (full agent minus agent without the translator). Orange: the
translator's own effect $A-\Sm$, against a read-only substitute $\Sm$ that
shows the same schema information. Blue: the effect of restoring that
information, $\Sm-C$. Percentage points, mean of three runs, with 95\% task
bootstrap intervals. Above the dotted line: original schemas; below: renamed
schemas (version~2). The ablation effect varies from null to large, while the
translator's own effect is negative in every setting.}
\label{fig:decomp}
\end{figure}

Table~\ref{tab:sr} gives success rates, Table~\ref{tab:decomp} the
decomposition and Fig.~\ref{fig:decomp} its two main components. On SQLite, Gemini's ablation effect is small and not
significant ($A-C=+3.2$\pp, $p=0.27$). Read alone, it says the translator
barely matters. The decomposition shows two significant effects of opposite
sign: with the same schema information, removing the translator
\emph{raises} success by $8.4$\pp{} ($A-\Sm$), and giving the agent physical
names, compared with having none, raises it by $11.6$\pp{} ($\Sm-C$).
Ablation reports their small difference.

BigQuery magnifies both effects. The ablation effect is still
non-significant ($+6.6$\pp, $p=0.07$), while the translator's own
contribution is $-18.3$\pp{} and the perception effect $+24.9$\pp. All three
substitutes beat $A$ on BigQuery (by $11.8$ to $18.3$\pp), and so does
$A^{+}$, which keeps the translator but adds a catalog tool ($+11.0$\pp):
on BigQuery, schema information that the translator does not provide is worth
a great deal, and the translator's SQL is a poor way to get it. On SQLite
$A^{+}$ and $A$ are equal ($A^{+}-A=-0.2$\pp), and the static schema of $S_1$ is
between $A$ and the tool-based substitutes on both databases.

The richer substitute $S_2$ does not explain the gap. $\Sm$, which shows
only what the translator shows, is at least as good as $S_2$ on both
databases ($+2.0$\pp{} on SQLite, n.s.; $+4.9$\pp{} on BigQuery,
$p=0.013$). The translator's cost is therefore not an artefact of a
substitute that sees more.

\noindent\textbf{Answer to RQ1.} Yes. Ablation alone would conclude that the
translator matters little; the decomposition shows that its operation hurts
and the view it provides helps, by similar amounts.

\subsection{Generality Across Backbones (RQ2)}
\label{subsec:rq2}
Qwen3.5-27B on SQLite has the same signs: the matched substitute is best
($64.2\%$), the translator's own contribution is negative ($-4.7$\pp, not
significant) and the perception effect positive ($+12.6$\pp, $p=0.0002$).
Unlike Gemini, its ablation effect is significant ($+7.9$\pp), so ablation
would at least detect a dependence; it would still attribute it to the
translator rather than to the view. On BigQuery the same model shows how
far that attribution can go wrong. Ablation now reports a large and highly
significant dependence ($A-C=+35.4$\pp, $p=10^{-4}$), because $C$ collapses to
$9.6\%$: without physical names, the agent executes a query in only 32--39 of
197 tasks per run and reaches the turn cap in $115.7$ task-runs per run. Yet
the matched substitute again beats $A$ ($57.9\%$ vs.\ $45.0\%$; $A-\Sm=-12.9$\pp,
$p=0.0004$), and $\Sm$ and $S_2$ are equal ($+1.2$\pp, $p=0.70$). The whole
ablation effect, and more, is the perception effect ($\Sm-C=+48.2$\pp). On the
original schema, then, a backbone that cannot infer physical names on its own
reproduces what renaming does to Gemini (Section~\ref{subsec:rq3}).

Qwen3.5-9B succeeds far less often ($15.6\%$ in $A$), and the ablation effect
is again null ($-1.2$\pp). Here the richer substitute is best: $S_2$ beats
$A$ by $11.4$\pp{} ($p=0.001$) and $\Sm$ by $6.2$\pp{} ($p=0.08$). A weak model
benefits from being shown whole models rather than the columns it asked for,
so for weaker backbones the content of the substitute matters, not only its
existence. The strong backbones show the opposite ordering ($\Sm\ge S_2$),
consistent with the extra columns acting as distraction once the model can
ask for what it needs.

\noindent\textbf{Answer to RQ2.} The signs of the two components (translator's
own contribution $\le 0$, perception effect positive) hold on a different
model family and at two sizes, although for the Qwen models not every
component is significant. The ablation verdict varies far more than its
components: null for Qwen3.5-9B, moderate for Qwen3.5-27B on SQLite and large
on BigQuery, where it credits the translator with what is in fact a perception
effect. The best substitute depends on backbone strength.

\subsection{Robustness to Renamed Schemas (RQ3)}
\label{subsec:rq3}
\begin{figure}[t]
\centering
\includegraphics[width=0.68\linewidth]{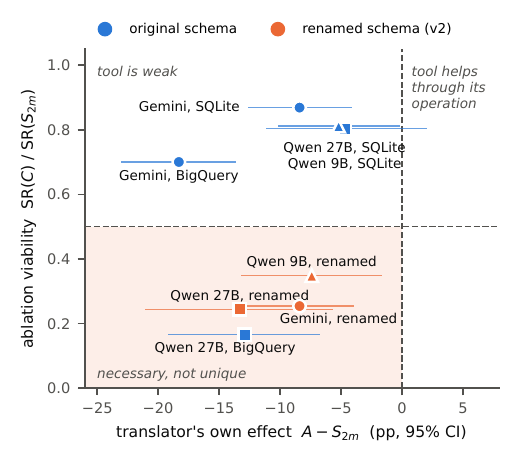}
\caption{Placement of each setting. Horizontal: the translator's own effect
with information held equal. Vertical: how much of the matched substitute's
success the ablation keeps. Renaming moves every backbone from ``tool is
weak'' to ``necessary, not unique'' while the translator's own effect stays
negative. Qwen3.5-27B on BigQuery lands there without renaming.}
\label{fig:placement}
\end{figure}

\begin{table}[t]
\caption{Renamed schemas, SR (\%), mean of three runs.\label{tab:renamed}}
\centering
\small
\begin{tabular}{@{}llrrrr@{}}
\toprule
Backbone & Schema & $A$ & $S_2$ & $\Sm$ & $C$ \\
\midrule
Gemini & original & 80.0 & 86.4 & 88.4 & 76.8 \\
       & renamed v1 & 69.1 & 84.0 & 84.4 & 0.0 \\
       & renamed v2 & 79.0 & 84.4 & 87.4 & 22.2 \\
\midrule
Qwen 27B & original & 59.5 & 56.8 & 64.2 & 51.6 \\
         & renamed v1 & 35.8 & 45.7 & 37.5 & 0.0 \\
         & renamed v2 & 47.4 & 57.5 & 60.7 & 14.8 \\
\midrule
Qwen 9B & original & 15.6 & 26.9 & 20.7 & 16.8 \\
        & renamed v1 & 5.2 & 21.5 & 10.9 & 0.0 \\
        & renamed v2 & 10.4 & 20.5 & 17.8 & 6.2 \\
\bottomrule
\end{tabular}
\end{table}

Renaming removes the agent's ability to guess physical names. $C$ falls from
$76.8\%$ to $22.2\%$ on Gemini, and from $51.6\%$ to $14.8\%$ on
Qwen3.5-27B (version~2; Table~\ref{tab:renamed}). The ablation effect
becomes very large ($+56.8$ and $+32.6$\pp): ablation alone would call the
translator indispensable. Yet the substitutes are almost unaffected
(Gemini $\Sm$: $88.4\%\rightarrow87.4\%$), and the information-matched
substitute still outperforms the translator, by $8.4$\pp{} on Gemini and
$13.3$\pp{} on Qwen3.5-27B, both significant. The translator moved from
``weak'' to ``necessary, not unique'' (Table~\ref{tab:quadrants},
Fig.~\ref{fig:placement}) without its
operation becoming more useful; only the scarcity of what it shows changed.

The residual $C$ success in version~2 comes from sample SQL inside layer
descriptions, which version~2 rewrote to the new names. On the 24 tasks whose
model list shows no renamed name, $C$ solves 0 of 72 task-runs on every
backbone. In version~1, where descriptions still use old names, $C$ is
$0\%$ on all three backbones, and $A$ loses $10.9$\pp{} on Gemini, about half
of it from SQL written with names taken from the outdated descriptions.

\noindent\textbf{Answer to RQ3.} The decomposition is not an artefact of
mirrored or memorized names. Removing that shortcut is exactly what turns
ablation's verdict around, while the translator's own contribution stays
negative.

\section{Analysis}
\label{sec:analysis}

\subsection{Is the Substitute Information-Matched?}
\label{subsec:info}
Table~\ref{tab:info} reports, for Gemini on SQLite, what each condition
showed the agent before its final query (Section~\ref{subsec:equiv}). $\Sm$
matches $A$: $19.7$ visible physical columns on average in both, and gold-column
coverage of $92\%$ and $90\%$. $S_2$ shows about half as many columns again
and $S_1$ more than twice as many. $C$ sees almost as many columns as $A$
($18.9$) because on the original schema semantic names mirror physical ones;
its lower gold coverage ($80\%$) is the information that only the
translator, or a substitute, supplies, and it is the reason renaming matters.

\begin{table}[t]
\caption{What each condition exposes (Gemini, SQLite, mean of three runs).
Columns: physical columns visible before the final SQL; share of the
database's columns; share of gold-query columns visible; prompt tokens per
task; tool calls per task.\label{tab:info}}
\centering
\small
\begin{tabular}{@{}lrrrrrr@{}}
\toprule
Cond. & $|V|$ & of DB & gold cov. & tokens & calls & SR \\
\midrule
$A$ & 19.7 & 33\% & 90\% & 40.9K & 4.0 & 80.0 \\
$A^{+}$ & 33.2 & 49\% & 96\% & 40.3K & 3.9 & 79.8 \\
$S_1$ & 44.3 & 67\% & 97\% & 11.5K & 1.3 & 84.0 \\
$S_2$ & 28.4 & 45\% & 96\% & 12.4K & 2.3 & 86.4 \\
$\Sm$ & 19.7 & 33\% & 92\% & 24.7K & 3.4 & 88.4 \\
$C$ & 18.9 & 32\% & 80\% & 18.0K & 2.3 & 76.8 \\
\bottomrule
\end{tabular}
\end{table}

Two other interface differences are visible in the table. First, cost:
$A$ uses about $3.3\times$ the prompt tokens of $S_2$ and $1.7\times$ those of
$\Sm$ for a lower success rate. Second, number of turns: $\Sm$ uses almost as
many tool calls as $A$, so its advantage is not a matter of shorter
episodes. Information is held equal on average; the interface is not, and we
read $A-\Sm$ as the effect of the translator's interface (composing SQL
through a structured query language) given equal information.

\begin{table}[t]
\caption{What each condition exposes on BigQuery, measured against the
semantic layer (29.6 physical columns per task on average). Mean of three
runs. Gold coverage on the 137 tasks with official
gold SQL.\label{tab:infobq}}
\centering
\small
\begin{tabular}{@{}lrrr@{\hspace{12pt}}rrr@{}}
\toprule
 & \multicolumn{3}{c}{Gemini 3.1 Pro} & \multicolumn{3}{c}{Qwen3.5-27B} \\
\cmidrule(lr){2-4}\cmidrule(l){5-7}
Cond. & $|V|$ & gold cov. & SR & $|V|$ & gold cov. & SR \\
\midrule
$A$ & 13.6 & 92\% & 64.8 & 13.4 & 92\% & 45.0 \\
$A^{+}$ & 17.5 & 97\% & 75.8 & -- & -- & -- \\
$S_1$ & 29.6 & 100\% & 76.6 & -- & -- & -- \\
$S_2$ & 20.0 & 99\% & 78.2 & 19.9 & 99\% & 56.7 \\
$\Sm$ & 13.7 & 95\% & 83.1 & 13.6 & 94\% & 57.9 \\
$C$ & 12.2 & 75\% & 58.2 & 11.5 & 71\% & 9.6 \\
\bottomrule
\end{tabular}
\end{table}

\noindent\textbf{BigQuery, and matching per task-run.} On BigQuery we
measure against the semantic layer rather than the full catalog: $V_X$
counts the physical columns in the layer's expressions
(Table~\ref{tab:infobq}). The averages match again: $|V|$ is $13.6$ for $A$
and $13.7$ for $\Sm$ with Gemini ($13.4$ and $13.6$ with Qwen3.5-27B), and
gold-column coverage is $92\%$ and $95\%$ ($92\%$ and $94\%$). The averages
hide a minority of task-runs in which $A$ saw fewer of the gold columns than
$\Sm$: 35 of 411 paired task-runs for Gemini and 52 for Qwen3.5-27B (27 of
399 on SQLite). These often follow a failed translator call (23 of 35 and 23
of 52, against about 30\% of the other task-runs), and in them $A$ trails
$\Sm$ by $46$ and $42$\pp. Where $A$ saw at least as much as
$\Sm$, $A-\Sm$ is $-13.3$\pp{} for Gemini and $-6.7$\pp{} for Qwen3.5-27B,
against $-16.1$ and $-11.2$\pp{} over all task-runs with gold SQL ($-5.4$
against $-8.5$\pp{} on SQLite). Most of the translator's cost therefore
remains when $A$ did see the information; the rest is the translator failing
to deliver what $\Sm$ returns by construction, which we count as part of its
interface. Because visibility depends on what the agent did, this split shows
co-occurrence, not cause.

\subsection{Where the Translator Loses}
\label{subsec:failures}
Three mechanisms account for most of the translator's negative own
contribution on SQLite; BigQuery adds a fourth, metrics that do not compile
(end of this subsection).

\noindent\textbf{Protocol failures.} The structured query of \csql{} is a
nested object, and Gemini sometimes answers with a native function call
instead of the text protocol. Even with three attempts, $A$ loses $5.0$ tasks
per run this way on SQLite ($12.3$ on renamed v1), against none for the
substitutes. Excluding these task-runs narrows $A-\Sm$ on SQLite from $-8.4$
to about $-5.3$\pp; the remainder is not protocol failure.

\noindent\textbf{Turn exhaustion.} Because the engine cannot express every
query, $A$ iterates over partial compilations and then writes SQL by hand.
Qwen3.5-27B reaches the 20-call cap in $14.3$ task-runs per run in $A$ on
SQLite, $29.0$ on renamed v2 and $49.3$ on BigQuery, against $0.7$--$4.3$
for $S_2$ and $\Sm$.

\noindent\textbf{Long hand-written SQL.} Conversely, the substitutes rely on
the agent writing the whole query. Qwen3.5-27B frequently omitted the final
brace of a tool call carrying long SQL; recorded as an invalid call, the
same reply was repeated until the turn cap. Before we repaired this in the
parser, $S_2$ and $\Sm$ lost 22--51 task-runs per run and the 27B ordering
appeared reversed ($A>S_2>\Sm$). After the repair, format loops are at most
$1.3$ per run and the ordering matches the other backbones. The repair applies
to all conditions; Gemini had no such failures and Qwen3.5-9B at most 12 of
405 task-runs, so their runs were not repeated.

On renamed v2, Qwen3.5-27B's $C$ loses $13.7$ task-runs per run without an
answer; 35 of these 41 fail because the conversation exceeds the model's context while the agent explores an
unfamiliar schema; they recur on the same tasks in every run and we count them
as failures of $C$.

\begin{table}[t]
\caption{BigQuery: task-runs per run (mean of three) that $A$ fails and
$\Sm$ solves, by cause. Each lost task-run gets the first cause that
applies, in the order listed.\label{tab:bqlosses}}
\centering
\small
\begin{tabular}{@{}lrr@{}}
\toprule
Cause & Gemini & Qwen 27B \\
\midrule
Infrastructure / protocol & 4.7 & 3.7 \\
Turn cap (20 calls) & 0.0 & \textbf{25.3} \\
Requested a metric that does not compile & 11.0 & 3.3 \\
Final SQL builds on translator output & 0.0 & 0.3 \\
Final SQL hand-written, translator worked & \textbf{25.7} & 8.3 \\
Translator never succeeded & 0.3 & 0.0 \\
\midrule
Total lost to $\Sm$ & 41.7 & 41.0 \\
Reverse ($A$ solves, $\Sm$ fails) & 5.7 & 15.7 \\
\bottomrule
\end{tabular}
\end{table}

\noindent\textbf{Losses on BigQuery.} Per run, Gemini's $A$ fails
$41.7$ tasks that $\Sm$ solves and solves $5.7$ that $\Sm$ fails;
Qwen3.5-27B's $A$ $41.0$ and $15.7$. Table~\ref{tab:bqlosses} assigns each
lost task-run one cause. \emph{Metrics that do not compile:} 67 of the 436
ported metrics fail to compile or dry-run on BigQuery
(Section~\ref{sec:testbed}). Gemini's $A$ asks the translator for one of them in
$48$ task-runs per run, and $11.0$ of its $41.7$ losses end in a wrong answer
after such a request ($3.3$ of $41.0$ for Qwen3.5-27B). \emph{Hand-written
SQL:} the translator previews its SQL against the single table a model is
mapped to, with \texttt{LIMIT 5}, and the agent almost never builds on it:
only $10$ (Gemini) and $11$ (Qwen3.5-27B) of $197$ task-runs per run reuse a
translator output in the final SQL, and these are mostly correct ($97\%$ and
$82\%$). The largest group of Gemini's losses, $25.7$ per run, are queries
written by hand after successful translator calls; in $43$ of these $77$
task-runs (three runs) at least one translator call had failed first. Losses
are over-represented on the four date-sharded databases whose models map to
one shard (Firebase, GA360, GA4, NOAA): $14.7$ of Gemini's $41.7$, against
$25\%$ of tasks. \emph{Turn exhaustion} dominates for Qwen3.5-27B: $25.3$ of
its $41.0$ losses hit the 20-call cap. Infrastructure and protocol failures
account for $4.7$ (Gemini) and $3.7$ (Qwen3.5-27B). On BigQuery, then, the
translator's cost is mainly that the agent spends turns on it and still
writes the answer itself; broken metrics explain about a quarter of Gemini's
losses, not most of them.

\subsection{Prompt Sensitivity}
\label{subsec:prompt}
The prompts of $A$ and $\Sm$ differ, and $A$'s prompt in particular coaches
the agent: its detailed instructions and worked examples tell it to use
\csql{} for exploration and to write the final query by hand. To test whether
the translator's negative own contribution is an artefact of this guidance,
we reran $A$ and $\Sm$ on SQLite with Gemini, three runs each, with
\emph{naive} prompts that keep the system instructions, the tool descriptions
and the response format but omit the detailed instructions and worked
examples ($A$: 1,320 to 530 words; $\Sm$: 559 to 381 words).

\begin{table}[t]
\caption{Prompt Sensitivity (Gemini, SQLite, Mean of Three Runs). Paired
Differences With 95\% Task Bootstrap Interval and Permutation $p$.\label{tab:prompt}}
\centering
\small
\begin{tabular}{@{}lrrr@{}}
\toprule
Prompt & SR($A$) & SR($\Sm$) & $A-\Sm$ \\
\midrule
Original & 80.0 & 88.4 & $-8.4$ {\scriptsize[$-12.6$, $-4.2$], .0002} \\
Naive & 76.0 & 88.4 & $-12.3$ {\scriptsize[$-17.0$, $-7.9$], .0001} \\
\midrule
Naive $-$ original & $-4.0$ {\scriptsize .08} & $0.0$ {\scriptsize 1.0} & \\
\bottomrule
\end{tabular}
\end{table}

Table~\ref{tab:prompt} shows that the guidance helps the translator, not the
substitute. Without it, $A$ loses $4.0$\pp{} ($p=0.08$) while $\Sm$ is
unchanged ($88.4\%$ in both), so the translator's own contribution grows from
$-8.4$ to $-12.3$\pp. Protocol failures do not explain the change: the naive
$A$ loses $3.0$ task-runs per run to them, against $5.0$ with the original
prompt, and with those task-runs left out the gap is $-10.6$\pp{} (naive)
against $-5.3$\pp{} (original). The negative own contribution is therefore
not produced by the prompt; if anything, the original prompt understates it.

\subsection{Case Study}
\label{subsec:case}
\begin{figure}[t]
\centering
\includegraphics[width=\linewidth]{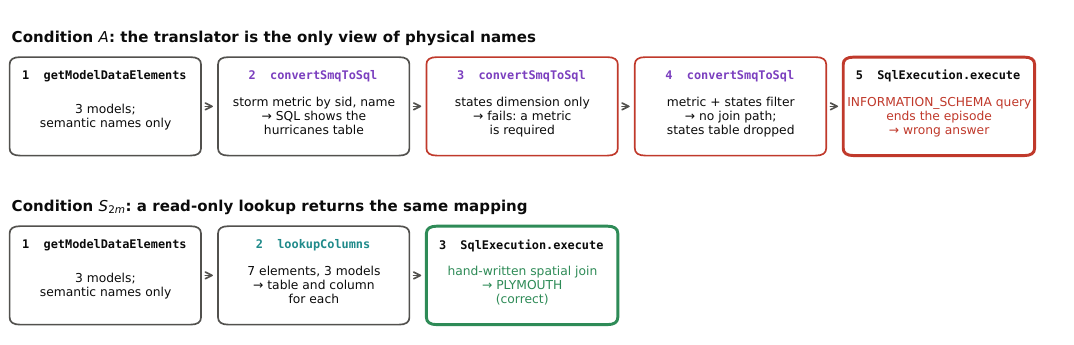}
\caption{Case study: BigQuery task bq277 with Gemini (run~1). The question
asks which port in a given World Port Index region, inside a U.S.\ state,
lies most often inside the wind radius of named North Atlantic tropical
storms. Both conditions see the same three semantic models. Under $A$ the
agent can learn physical names only through \csql{}, whose engine needs a
metric and joins one hop, so the states table never appears; the agent falls
back to a catalog query, which ends the episode. Under $\Sm$ one
\lcol{} call returns the table and column of every element, and the agent
writes the spatial join itself. $A$ fails this task in all three runs and
$\Sm$ solves it in all three.}
\label{fig:case}
\end{figure}

Fig.~\ref{fig:case} follows one task through $A$ and $\Sm$. The task needs
three tables from three semantic models with no join path between them in
the layer. Under $A$, the first \csql{} call reveals the hurricanes table;
the second, a states dimension without a metric, is rejected; the third adds
a metric, and the engine silently drops the states filter because no join
path connects the models. Having seen only one of the three tables, the agent
queries the catalog through \sqlex{}, which ends the episode with a list of
table names. Under $\Sm$ the agent asks \lcol{} for seven elements across the
three models, receives each one's table and column, and writes the spatial
join in one step. The information the agent needed is the same in both
conditions; what differs is whether it can be obtained without composing a
query the engine can compile. This is the mechanism behind the
``hand-written'' and ``metric does not compile'' rows of
Table~\ref{tab:bqlosses}.

\subsection{Noise Floor}
\label{subsec:noise}
Repeated runs of $A$ differ by at most $2.2$\pp{} in success rate on SQLite
(Gemini), but $19$--$20\%$ of tasks change outcome between any two runs and
$30\%$ differ across the three; on BigQuery the run-to-run differences are
$1.0$--$6.1$\pp{} with $18$--$22\%$ of tasks flipping. Single-run comparisons of
a few points are therefore within noise, which is why every main comparison
averages three runs per task before testing. The components we highlight
($8$--$25$\pp) exceed this floor; the ablation effects we call
non-significant do not.

\subsection{Computational Cost}
\label{subsec:cost}
The decomposition adds one condition to an ablation study, and that condition
is cheaper to run than the full system. On SQLite with Gemini, $\Sm$ uses
$24.7$K prompt tokens and $3.4$ tool calls per task against $40.9$K and $4.0$
for $A$ (Table~\ref{tab:info}). All Gemini BigQuery runs together (six
conditions, three runs each) issued about 10,300 query jobs and billed
8.2\,TB, about 51\,USD at on-demand pricing; no query reached the 200\,GB
per-query cap we set. The substitutes themselves cost nothing to build by
hand: one script generates them from the catalog and the semantic layer.

\section{Discussion and Limitations}
\label{sec:discussion}

\subsection{Discussion}
\noindent\textbf{What ablation measures.} Ablation is not wrong: $A-C$ is the
dependence of this configuration on the tool package, and in our testbed it
correctly says that removing the translator with nothing in its place costs
little on Spider for Gemini and a great deal once names are opaque. What it cannot say
is why. In our data the answer is the same in every setting: the
translator's operation contributes nothing or less than nothing, and its view
of the schema contributes everything, while the ablation verdict ranges
from ``irrelevant'' to ``indispensable''. A design decision based on the
ablation verdict (keep the translator because removing it collapses the
agent) would keep the wrong component.

\noindent\textbf{When to run the decomposition.} The extra condition is
needed when a tool returns information about the environment that the agent
cannot otherwise obtain. Such tools are common: schema and catalog tools,
document and page readers, file viewers, state inspectors. For these the
substitute is usually easy to build, because what the tool exposes is
already structured: in our case the static schema, \dsch{}, \lcol{} and the
ablation prompt are generated by a script from the catalog and the semantic
layer. Measuring informational equivalence needs only logs and an
enumeration of the relevant environment state.

\noindent\textbf{Choosing the substitute.} A substitute that exposes more
than the tool confounds $A-S$ with an information effect, and in our data
that effect changes sign with backbone strength ($\Sm-S_2$: $-6.2$\pp{} for
the 9B model, $+1.2$ to $+7.4$\pp{} for the stronger ones). An
information-matched substitute, verified by measurement, is therefore the
primary one; richer substitutes are useful as additional rungs. This mirrors
schema-linking studies, in which showing the model less of the schema can help
or hurt depending on the schema and the model
\cite{luoma2026schema,chung2025long}.

\noindent\textbf{Implications for semantic-layer agents.} The value of the
semantic layer for these agents lies in its vocabulary and its mapping to
physical columns, not in its query compiler: $\Sm$, which exposes only the
mapping, is the best condition for Gemini and Qwen3.5-27B on both dialects
and the second best for Qwen3.5-9B on SQLite, at about $60\%$ of $A$'s token
cost on SQLite (Gemini). For production
systems this suggests exposing the layer's mapping as a lookup tool and
letting the model write SQL, rather than routing queries through a compiler
whose expressiveness is narrower than the model's.

\subsection{Limitations}
\label{sec:limitations}
\noindent\textbf{One agent, one benchmark family.} All results come from
one semantic-layer text-to-SQL agent on one benchmark family,
Spider~2.0-Lite. We vary what can be varied within it: two database engines
with different dialects and schema scale, three backbones of different
strength, and a renamed-schema control that changes how much the physical
names reveal. The pattern of signs holds across all of them, but we have not
tested other text-to-SQL benchmarks or other domains. The decomposition
itself only requires a tool that both acts and exposes part of the
environment, and a read-only channel that can return the same information.
The finding that the translator's
own contribution is negative is a property of this tool and configuration,
not a general claim about tools.

\noindent\textbf{Public schemas.} Spider~2.0 databases are public. The
renamed control removes name mirroring and memorized names on SQLite but was
not applied to BigQuery, and data values themselves may be familiar.
Spider~2.0 gold annotations are also known to contain errors
\cite{jin2026pervasive}. Our comparisons are paired on the same gold results,
which removes errors common to all conditions from the differences, but not
errors whose effect depends on how a condition writes its query.

\noindent\textbf{Engine defects count against $A$.} The engine joins only one
hop, needs plain-column join keys and fails to compile about one metric in
ten (one in seven on BigQuery). We treat these as properties of the tool under
study; a better compiler would raise $A$, and the translator's own
contribution is specific to this engine.

\noindent\textbf{Interface differences.} Substitutes differ from $A$ in
prompt text, turn structure and token count. We measured information,
tokens and turns, and tested prompt guidance on one setting
(Section~\ref{subsec:prompt}), but not every aspect of presentation; $S_1$, which puts
the whole schema in the prompt, also tests sensitivity to prompt layout.

\noindent\textbf{Coverage.} Informational equivalence was measured for
Gemini on SQLite (Table~\ref{tab:info}) and for both backbones on BigQuery
(Table~\ref{tab:infobq}), where it is measured against the semantic layer
rather than the full catalog. For the Qwen backbones on SQLite and for the
renamed schemas, $\Sm$ is matched by construction but not by measurement. Qwen3.5-9B was not run on BigQuery. The
BigQuery $C$ condition must infer projects and datasets that the substitutes
receive. Version~2 of the renamed layers also rewrote ordinary English words
that coincided with old table names, which garbles some descriptions in every
condition. Gemini~3.1~Pro is a preview model and partly ignores a zero
thinking budget.

\noindent\textbf{Stale descriptions in the BigQuery layers.} The BigQuery
port rewrote the layers' expressions but not their descriptions: by a pattern
search, about 230 of the 3,022 descriptions, in 45
of the 60 ported layers and covering 148 of the 197 tasks, still contain
Snowflake-specific guidance, such as Snowflake table names or double-quoted
column names, which BigQuery reads as string literals. Every condition sees
these descriptions. Restricted to the 49 tasks whose layers contain no such
guidance, the decomposition keeps its signs and size: $A-\Sm$ is $-13.6$\pp{}
for Gemini ($p=0.009$) and $-13.6$\pp{} for Qwen3.5-27B ($p=0.08$), and
$\Sm-C$ is $+33.3$ and $+51.7$\pp.

\section{Conclusion}
\label{sec:conclusion}
Removing a tool removes what it does and what it shows. One extra
condition, a read-only substitute that restores what the tool showed, with
its informational match to the tool measured rather than assumed, splits the
ablation effect into the two. On a semantic-layer text-to-SQL agent across two SQL dialects, three
backbones and a renamed-schema control, the split changes the conclusion:
ablation is non-significant where two large opposing effects cancel, and
declares the tool indispensable where a substitute outperforms it. Tool
evaluation for agents whose tools double as perception should report the
decomposition alongside the ablation.

In future work, we plan to apply the decomposition outside databases. Page
readers in web agents and file viewers in coding agents also both act and
expose part of the environment, and their read-only substitutes can be
generated from the page or the file system in the same way. A second
direction is to measure informational equivalence per task rather than on
average, and to extend it from schema names to data values and join paths.
Finally, for semantic-layer systems, a lookup tool that returns the layer's
mapping could be combined with a stronger compiler, to test whether a better
translator can recover a positive contribution of its own.

\section*{Data and Code Availability}
The data and code that support the findings of this study are available from
the corresponding author upon reasonable request.

\section*{Acknowledgment}
This work was supported by the Technology Development Program
(RS-2026-25515163) funded by the Ministry of SMEs and Startups (MSS, Korea).

\clearpage
\bibliographystyle{IEEEtran}
{\small
\bibliography{refs}
}

\end{document}